\documentclass[letterpaper]{article} % DO NOT CHANGE THIS
\usepackage{aaai24}  % DO NOT CHANGE THIS
\usepackage{times}  % DO NOT CHANGE THIS
\usepackage{helvet}  % DO NOT CHANGE THIS
\usepackage{courier}  % DO NOT CHANGE THIS
\usepackage[hyphens]{url}  % DO NOT CHANGE THIS
\usepackage{graphicx} % DO NOT CHANGE THIS
\usepackage{natbib}  % DO NOT CHANGE THIS AND DO NOT ADD ANY OPTIONS TO IT
\usepackage{caption} % DO NOT CHANGE THIS AND DO NOT ADD ANY OPTIONS TO IT
\usepackage{multirow}
\usepackage{amsmath}
\usepackage{algorithmic}
\usepackage[caption=false]{subfig}
\usepackage{booktabs}
\usepackage[linesnumbered,ruled,vlined]{algorithm2e}
\SetKwInput{KwInput}{Input}
\SetKwInput{KwOutput}{Output}
\usepackage{newfloat}
\usepackage{listings}
\DeclareCaptionStyle{ruled}{labelfont=normalfont,labelsep=colon,strut=off} % DO NOT CHANGE THIS
\title{Elevating Defenses: Bridging Adversarial Training and Watermarking for 
\\ Model Resilience}
\author{
    Janvi Thakkar\textsuperscript{\rm 1}, Giulio Zizzo\textsuperscript{\rm 2}, Sergio Maffeis \textsuperscript{\rm 1}\\
}
\affiliations{
    \textsuperscript{\rm 1}Department of Computing, Imperial College London\\
    \textsuperscript{\rm 2} IBM Research Europe
    janvi.thakkar22@imperial.ac.uk, giulio.zizzo2@ibm.com, sergio.maffeis@imperial.ac.uk
}

\usepackage{bibentry}
\begin{document}

\maketitle

\begin{abstract}
Machine learning models are being used in an increasing number of critical applications; thus, securing their integrity and ownership is critical. Recent studies observed that adversarial training and watermarking have a conflicting interaction.
This work introduces a novel framework to integrate adversarial training with watermarking techniques to fortify against evasion attacks and provide confident model verification in case of intellectual property theft. We use adversarial training together with adversarial watermarks to train a robust watermarked model. The key intuition is to use a higher perturbation budget to generate adversarial watermarks compared to the budget used for adversarial training, thus avoiding conflict. We use the MNIST and Fashion-MNIST datasets to evaluate our proposed technique on various model stealing attacks. The results obtained consistently outperform the existing baseline in terms of robustness performance and further prove the resilience of this defense against pruning and fine-tuning removal attacks. 
\end{abstract}

\section{Introduction} 
Adversarial training is one of the most widely used techniques to defend against model evasion attacks. The process requires to train a model on adversarial examples; however, generating these examples is expensive. For instance, the \emph{projected gradient descent} approach~\cite{madry2017towards} uses an iterative optimization technique to generate adversarial examples, requiring high computational resources to train a robust model. When deployed, the computational value invested into the models makes them higher-valued targets, increasing the risk of model theft. Several digital watermarking (model watermarking) techniques were proposed in the literature to verify the ownership of the stolen machine learning (ML) model. However, the techniques concentrated on devising a watermarking strategy in the general setting for copyright protection, without considering the impact it might have when combined with other protection mechanisms. A recent study \cite{szyller2022conflicting}, which explored the impact of how different protection mechanisms interact, found that adversarial training and model watermarking, when combined, have conflicting interactions. While the watermarking accuracy remains high for the above combination, they observed that it decreases the robustness of the model toward evasion attacks. They attributed this to the use of out-of-distribution watermarks, which use ``distinct labels compared to the training dataset, making it easier for the evasion attack
to find a perturbation that leads to a misclassification''~\cite{szyller2022conflicting}.

In this work, we propose a novel way of combining existing defenses to get better robustness against evasion attacks, while maintaining the same watermarking performance. 

In particular, we do not want the watermarks to interfere with the goals of adversarial training.
Our key idea is to generate the watermarks themselves with adversarial training, so that they have a similar distribution to the training set, which already includes adversarial samples. Crucially though, we use a specific (higher) perturbation budget in the watermark generation, so that watermarks can still be distinguished for the purpose of ownership verification.

We tested our approach on the MNIST and Fashion-MNIST datasets. We empirically proved the effectiveness of our design by providing an in-depth analysis of performance for various model stealing and removal attacks.

The main contributions of our work are:
\begin{enumerate}
    \item We propose a novel way of embedding adversarial watermarks in an adversarially trained model. We empirically show that the proposed design is as robust as the non-watermarked robust model, and as efficient as the non-robust watermarked model compared to the baseline \cite{szyller2022conflicting}.
    \item We provide an in-depth analysis of the effectiveness of our approach for various model stealing scenarios, i.e., black-box, grey-box, and white-box model stealing attacks. Furthermore, we demonstrate the robustness of our approach under two types of removal attacks: fine-tuning and pruning attacks.
\end{enumerate}

\section{Related Work}
The notion of adversarial examples was introduced in \cite{szegedy2013intriguing}, where the researchers observed that applying a small imperceptible perturbation to the input image can alter the model predictions. In particular, the study \cite{goodfellow2014explaining} proposed the Fast Gradient Sign Method (FGSM), which uses a single-step gradient method to generate adversarial samples. The samples were generated and added to the training dataset before the normal training process, later known as adversarial training. Multiple works \cite{tramer2017ensemble,kurakin2018adversarial} enhanced and built upon the FGSM approach. These improvements led to the introduction of projected gradient descent (PGD), a widely used technique for training models using adversarial samples \cite{madry2017towards}. Owing to the widespread application of PGD, we use it as one of the primary algorithms in our research to defend against the evasion attack. 

The work \cite{uchida2017embedding} was the first to watermark the deep neural networks (DNNs), where they proposed inserting a unique key vector during the training process. In the case of intellectual property theft, one can verify the ownership of the model by checking for the inserted key vector in the matrix of model parameters. However, this approach required white-box access to verify the model, which is not always plausible in the real world. Another work \cite{le2020adversarial} proposes to use adversarial samples as watermarks in the black-box setting. They claim that the models with these watermarks can confidently verify their ownership owing to the high transferability of adversarial samples \cite{szegedy2013intriguing}. Many other recent works \cite{szyller2021dawn} \cite{szyller2022conflicting} use the backdoor watermarking technique, whose aim is to insert out-of-distribution (OOD) watermarks during the training process. This approach is adopted to ensure that the watermarks used are distinct from the main learning objective of our ML model. However, the study \cite{szyller2022conflicting} observed that combining OOD watermarks with adversarial training decreases the robustness of the model against evasion attacks. They reason that because the labels in the OOD dataset are distinct from the actual training set, it is simpler for an evasion attack to identify a perturbation that causes inaccurate predictions. 

Thus, in this work, we propose to use watermarks with a similar distribution to our adversarial training dataset to enhance the robustness. The idea is to use watermarks generated using adversarial training, similar to the work by \cite{le2020adversarial}. However, the study \cite{namba2019robust} debates that one cannot efficiently use the adversarial watermarks together with the model trained using adversarial training; \emph{``as this watermarking method may falsely determine the models without watermarks as models with watermark \cite{namba2019robust}."} Nonetheless, as detailed in the sections below, we argue that one \emph{can} still use the watermarks generated using adversarial training with a specific parameter setting without having any conflicting interaction.

\section{Approach}
% \subsection{Preliminaries}
% \textbf{OOD Dataset:}The OOD stands for out-of-distribution dataset. These data samples significantly differ from the training dataset on which our machine-learning model is trained.\\

% \textbf{Adversarial Samples:} These are the adversarial examples generated either using FGSM or PGD techniques. The perturbation budget for adversarial samples is the same as that used in adversarial training of our ML model.\\

% \textbf{Adversarial Watermarks:} These are the adversarial samples generated using an \textit{existing adversarially trained model} before the training process. As we use these adversarial samples to watermark our model, we call them \textit{adversarial watermarks}. The perturbation budget is decided based on our proposed technique, as defined in the section below. Please note that when we say \textit{existing adversarially trained model}, we mean to use a non-watermarked robust model to generate the watermarks and use these adversarial watermarks to train the combined models. 

\subsection{Baseline}
In this work, we use the study \cite{szyller2022conflicting} as our baseline to compare with our proposed combination of model watermarking and adversarial training. The key idea of the baseline approach was to use PGD-based adversarial training to defend against the model evasion attacks, and combine it with the backdoor watermarking technique to watermark the trained model for ownership verification. They proposed to use out-of-distribution (OOD) watermarks. The intuition was to select the watermarking set that is unique to the training dataset and does not conflict with the primary objective of model training.

Implementing the above approach is simple, as described in the pseudo-code in the Appendix: Algorithm~\ref{adv_water_algo}. The overall procedure can be summarised below:
\begin{enumerate}
    \item Adapt the standard adversarial training procedure, where the data points are perturbed (with PGD) based on the specified perturbation budget ($\beta$) at every iteration.
    \item The OOD dataset (watermarking set) is provided by the model owner in advance.
    \item During the training phase, the watermarking set can be added to the training set, or the model can be separately trained on it at the end of every epoch.
\end{enumerate}  
\textbf{Limitations:} Both adversarial training and the watermarking, when applied individually, work effectively for the purpose they were designed for. However, when applied simultaneously, they have a conflicting interaction. The study \cite{szyller2022conflicting} observed that baseline interaction has good performance with respect to the utility of the model; however, it affects the adversarial performance and decreases its robustness against evasion attacks. They attribute performance degradation as a result of using OOD watermarks, which use labels distinct from those in the actual training dataset, thus altering the model decision boundaries. As a result, this makes it easier for an evasion attack to identify a perturbation that causes incorrect results.

\subsection{Proposed Technique}
We observed on the baseline that using an OOD dataset for watermarking leads to a decrease in robustness performance. We hypothesize that this occurs due to overfitting the model on the watermarking dataset, which affects the models generalization ability by shifting its decision boundaries. Despite the shift in decision boundaries, the test performance remains unaffected as the watermarking dataset size is insufficient to significantly alter the overall behavior of the model. However, this drift in decision boundaries adversely affects the model robustness against adversarial samples. The adversarial samples are designed to exploit small vulnerabilities in the decision-making process. Even slight shifts in decision boundaries can make it easier to find samples that can compromise the robustness, leading to successful evasion attacks. 
\begin{figure}[h!]
\centering
\includegraphics[width = \linewidth]{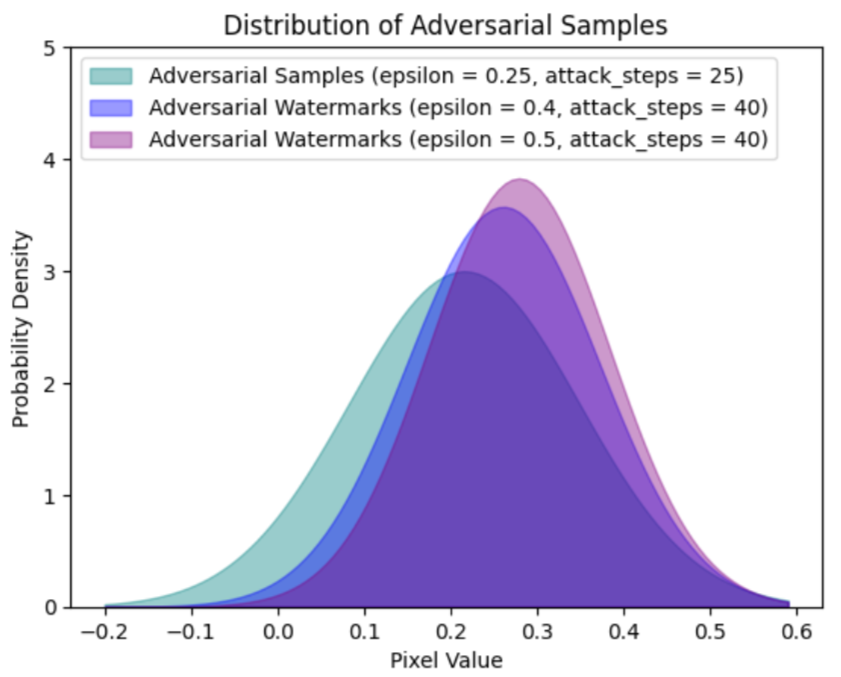}
\caption{The figure shows the distribution of adversarial samples and the adversarial watermarks, with different perturbation budgets.}
\label{fig:samples1}
\end{figure}

To counter this, we propose to use watermarks generated via adversarial training, also known as \textit{adversarial watermarks}. They are often used in the literature \cite{le2020adversarial} owing to their high transferability to stolen ML models. The idea is to use watermarks that are distinct compared to the training samples, but have a similar distribution to our adversarial training dataset. In our case, the training dataset comprises original data samples and their respective perturbed adversarial samples. We cannot use watermarks with the same distribution as the training set because it would be difficult to differentiate them, and may provide a false sense of verification. Instead, we hypothesise that watermarks generated using an adversarial training technique will have a similar distribution, i.e., they share certain statistical properties as that of adversarial samples in the training set. As shown in Appendix:~Figure \ref{fig:samples}, the distribution of the OOD and adversarial dataset differs significantly in the case of baseline, while in our proposal, watermarks have a similar distribution as that of the adversarial training set (Figure \ref{fig:samples1}). 

\begin{figure}[h!]
\centering
\includegraphics[width = 2in]{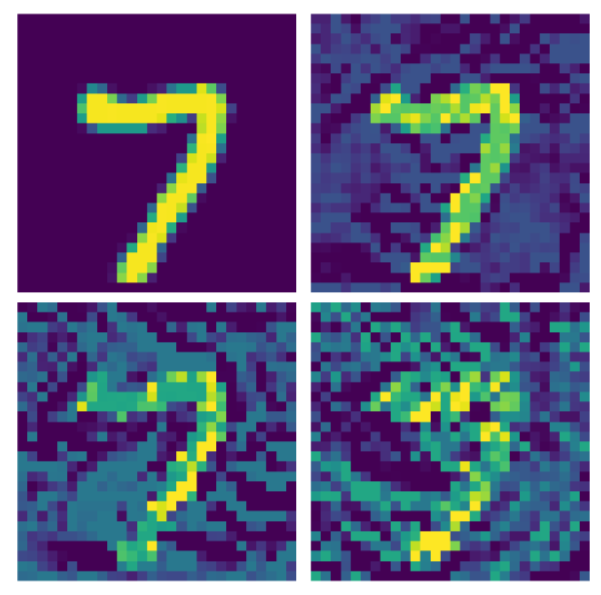}
\caption{The figure shows an instance of training data point (top left) from MNIST dataset, its corresponding adversarial sample (top right), and watermarks (bottom left with 0.4 epsilon budget, and bottom right with 0.5 epsilon budget)}
% Th figure on the right shows an instance of training data point (top left) from MNIST dataset, its corresponding adversarial sample (top right), and watermarks (bottom left with 0.4 epsilon budget, and bottom right with 0.5 epsilon budget)
\label{fig:samples1_un}
\end{figure}

However, one may wonder if adversarial samples and adversarial watermarks will be too similar, and conflict at inference time. In fact, adversarial samples and adversarial watermarks differ in the way they are crafted (Appendix:~Figure \ref{fig:samples1_un}). We propose to generate the watermarks using adversarial training with a higher perturbation budget than the adversarial samples. We claim that there exists a lower bound for an epsilon ($\epsilon$-perturbation budget) with which adversarial training can be used effectively, after which the utility of the model degrades, making it ineffective. We leverage this knowledge to generate adversarial watermarks with a higher perturbation budget to differentiate them from the adversarial samples. In addition, we empirically observed, as was also reported in \cite{madry2017towards}, that when we apply adversarial training to improve the robustness within some $\epsilon$-neighbourhood, it exhibits effectiveness for ($\epsilon+\alpha$)-neighborhood, where $\alpha$ is a positive constant. Thus, we use $\beta$ perturbation budget, where $\beta > \epsilon + \alpha$, to generate the adversarial watermarks.

The training process is similar to the baseline approach, but we substitute the OOD dataset with adversarial watermarks. These watermarks are derived from samples within the original training set, aligning with the distribution of adversarial samples, also generated from the training dataset. The main difference between adversarial samples and watermarks lies in the perturbation budget employed during their generation. While they exhibit similar statistical properties, they also remain unique. This uniqueness is crucial to avoid confusion between watermarked models and those influenced solely by adversarial samples. Moreover, adversarial watermarks need to be generated before the training procedure using an existing adversarially trained model. This is done to ensure that watermarks possess the desired adversarial characteristics already learned by the model, making them robust and difficult to remove. 

\section{Experimental Setting}
\subsection{Dataset, Model and Metric} 
We used MNIST \cite{deng2012mnist}, and the Fashion-MNIST (FMNIST) \cite{xiao2017fashion} datasets to compare our proposed technique with the baseline approach. For both the datasets, we use a 4-layer convolutional model comprising two consecutive convolutional blocks followed by two linear layers. The hyperparameters used for the experiments are defined in Table~\ref{hyper-params} in the Appendix.
Please note that we used FMNIST dataset as OOD for MNIST, and vice-versa, however, there is no direct correlation. Any arbitrary data points can be chosen as watermarks provided they are out-of-distribution with respect to the training dataset. 

We measure our model performance through test accuracy on a standard test dataset, robustness accuracy on adversarial test samples, and watermarking accuracy on the OOD/adversarial watermarking set.

\begin{table*}[h!]
\centering
\begin{tabular}{ c c cc cc cc }
\toprule
\multirow{2}{*}{\textbf{Dataset}} &
  \textbf{No def.} &
  \multicolumn{2}{c }{\textbf{ADVTR}} &
  \multicolumn{2}{c }{\textbf{WM (OOD)}} &
  \multicolumn{2}{c }{\textbf{WM (Adversarial)}} \\
 &
  \textbf{Test Acc} &
  \multicolumn{1}{c }{\textbf{Test Acc}} &
  \textbf{Adv Acc} &
  \multicolumn{1}{c }{\textbf{Test Acc}} &
  \textbf{Water Acc} &
  \multicolumn{1}{c }{\textbf{Test Acc}} &
  \textbf{Water Acc} \\ \midrule
\textbf{MNIST}  & \multicolumn{1}{|c|}{99.57} & \multicolumn{1}{|c|}{99.2}  & 92.82 & \multicolumn{1}{|c|}{99.36} & 100 & \multicolumn{1}{|c|}{99.45} & 100 \\
\textbf{FMNIST} & \multicolumn{1}{|c|}{93.64} & \multicolumn{1}{|c|}{86.91} & 70.95 & \multicolumn{1}{|c|}{90.96} & 100 & \multicolumn{1}{|c|}{92.29} & 99  \\ \bottomrule
\end{tabular}
\caption{Performance of the model with different strategies, with and without defense mechanisms for MNIST and Fashion-MNIST dataset.}
\label{tab:performance1}
\end{table*}

% \usepackage{multirow}
\begin{table*}[h!]
\centering
\begin{tabular}{c ccc ccc}
\toprule
                & \multicolumn{3}{c}{\textbf{ADVTR + WM (OOD)}}          & \multicolumn{3}{c}{\textbf{ADVTR + WM (Adversarial)}} \\
\multirow{-2}{*}{\textbf{Dataset}} &
  \multicolumn{1}{c}{\textbf{Test Acc}} &
  \multicolumn{1}{c}{\textbf{Adv Acc}} &
  \textbf{Water Acc} &
  \multicolumn{1}{c}{\textbf{Test Acc}} &
  \multicolumn{1}{c}{\textbf{Adv Acc}} &
  \textbf{Water Acc} \\ \midrule
\textbf{MNIST} &
  \multicolumn{1}{|c|}{99.02} &
  \multicolumn{1}{c|}{88.39} &
  100 &
  \multicolumn{1}{|c|}{99.03} &
  \multicolumn{1}{c|}{92.01} &
  100 \\
\textbf{FMNIST} & \multicolumn{1}{|c|}{85.42} & \multicolumn{1}{c|}{57.75} & 100 & \multicolumn{1}{|c|}{86.49} & \multicolumn{1}{c|}{65.84} & 93 \\ \bottomrule
\end{tabular}
\caption{Performance of the model with simultaneous deployment of adversarial training and model watermarking technique while using OOD and adversarial watermarks.}
\label{tab:performance2}
\end{table*}
\subsection{Model Stealing Techniques}
In addition to comparing the effectiveness of the approach using accuracy, we evaluate the technique by launching different model stealing attacks. The objective is to see how robust our inserted watermarks are in verifying the ownership of the model in the case of intellectual property theft. In this experimental setup, we assume that we only have black-box access to the stolen model, and it can be queried to see how it performs on our watermarking dataset. If it is above some threshold, we can successfully verify our ownership. In this work, we conclude that the watermarking strategy is successful if we can reliably predict more than 50\% of watermarking samples, i.e., if the transferability rate of watermarks ranges from 50\% and above. Furthermore, based on the information one has about the model they can launch various model stealing attacks:
\begin{enumerate}
    \item \textbf{Black-box setting:} In this particular setting, our adversary ($\mathcal{A}$) has no direct access to the trained model or any other internal working. However, they can query the API to gain information about its performance of various inputs. For each query, we only output the class label of the input image predicted by our model and do not provide any information about the class logits. Finally, our $\mathcal{A}$ uses the information about the queried input-output pairs to train a duplicate model that has a identical test performance as that of our original model. 
    \item \textbf{Grey-box setting:} In this setting, our $\mathcal{A}$ has partial information about the way our model was trained. Specifically, for our setup, we assume that $\mathcal{A}$ knows the detailed architecture of the model and the fact that it was trained using an adversarial training process. We use a similar setting, as described in the black-box setting, to steal the model by querying the API to gain information and consequently train the duplicate model. 
    \item \textbf{White-box setting:} In this setting, our $\mathcal{A}$ has direct access to the model. For instance, the cloud provider where our model was deployed, is malicious, and it redistributed the model to some 3rd party, thus giving full access to our original model and its workings. 
\end{enumerate}
To verify the stolen model, one only needs black-box access. Let us say the stolen model is deployed on some cloud platform, and we suspect it to be our model; then, we can simply query the stolen model to see how it performs on our watermarking dataset. The performance can be a good first indicator before taking any legal action against the malicious adversary. Thus, it becomes crucial that our watermarking technique has a high transferability property in case of various model stealing attacks.

\subsection{Removal Attack: Pruning and Fine-tuning}
Even after stealing a model using the above setting, an adversary can also try to remove the watermarks by applying various removal attacks. In this work, we investigate the robustness of our watermarks on two removal attacks: fine-tuning and pruning attacks. For fine-tuning the attack, we further train the stolen model on the main classification task with additional training data for 40 epochs. During fine-tuning, all layers are fine-tuned, and their weights are updated. For a pruning attack, we prune our stolen model with a range of pruning rates, starting from 10\% to 90\% of the parameters. To accomplish this, we employ the $L_1$ unstructured pruning \cite{zhu2017prune} technique to prune our model parameters.
\section{Results and Analysis}
This section provides a detailed analysis of the two approaches. However, before diving into the performance, we want our readers to know the rationale behind selecting the specific perturbation budget used to generate adversarial watermarks. As proposed, there exists a perturbation budget $\beta$, which can be successfully used to generate the adversarial watermarks without them being in conflict with the adversarial samples. Thus, we initially trained models for both datasets, using adversarial training with a pre-defined perturbation budget (MNIST - ($\epsilon = 0.25$, num. steps $= 25$) and FMNIST - ($\epsilon = 0.15$, num. steps $ = 15$)). The pre-defined budgets were selected based on their utility at the inference time. We observed that adversarial samples generated using a higher perturbation budget have perceptible noise and thus lose their utility to be used in evasion attacks. We further observed (from Appendix~Table \ref{only-adv-1}) that when this adversarially trained model is used to detect adversarial samples generated with higher perturbation budget, they are marginally robust to some larger budget ($\epsilon + \alpha$), but have a poor performance with any increase in epsilon above that. Thus, we leverage these higher perturbation budget values to generate the adversarial watermarks that have a similar distribution and have no conflict with adversarial samples generated during the training process. We use a perturbation budget of 0.4 and 40 attack steps for the MNIST dataset and a 0.3 perturbation budget and 40 attack steps for the FMNIST dataset.
% \usepackage{multirow}
\begin{table*}[h!]
\centering
\begin{tabular}{c ccc ccc}
\toprule
\multicolumn{7}{c}{\textbf{Black-box Transferability}} \\ \toprule
\multicolumn{1}{c}{} &
  \multicolumn{3}{c}{\textbf{AdvTraining + WM (OOD)}} &
  \multicolumn{3}{c}{\textbf{AdvTraining + WM (Adversarial)}} \\
\multicolumn{1}{c}{\multirow{-2}{*}{\textbf{Dataset}}} &
  \multicolumn{1}{c}{\textbf{Test Acc}} &
  \multicolumn{1}{c}{\textbf{Adv Acc}} &
  \multicolumn{1}{c}{\textbf{Water Acc}} &
  \multicolumn{1}{c}{\textbf{Test Acc}} &
  \multicolumn{1}{c}{\textbf{Adv Acc}} &
  \textbf{Water Acc} \\ \midrule
\multicolumn{1}{c|}{\textbf{MNIST}} &
  \multicolumn{1}{c|}{89.24} &
  \multicolumn{1}{c|}{1.29} &
  \multicolumn{1}{c|}{6} &
  \multicolumn{1}{c|}{88.36} &
  \multicolumn{1}{c|}{0.33} &
  54 \\
\multicolumn{1}{c|}{\textbf{FMNIST}} &
  \multicolumn{1}{c|}{72.47} &
  \multicolumn{1}{c|}{3.96} &
  \multicolumn{1}{c|}{8} &
  \multicolumn{1}{c|}{73.86} &
  \multicolumn{1}{c|}{10.09} &
  56 \\ \bottomrule
\end{tabular}
\caption{Transferability of performance for various metric when the models undergo black-box model stealing attack.}
\label{tab:black-box-performance}
\end{table*}

% \usepackage{multirow}

\subsection{Performance}
The performance of our model with different defense strategies is presented in Table \ref{tab:performance1}. We can observe that all the strategies perform well on the task they were designed for. Furthermore, Table \ref{tab:performance2} illustrates the performance of the combined effect of deploying adversarial training and watermarking techniques. We can observe that for both the approaches, the baseline (OOD watermarks), and our proposed strategy (adversarial watermarks) perform strongly in terms of the model utility and watermark verification. However, the adversarial accuracy when trained using OOD watermarks witnesses a drop of around 4\% (from 92.82\% $\rightarrow$ 88.39\%) for the MNIST dataset and around 13\% drop (from 70.95\% $\rightarrow$ 57.75\%) for the FMNIST dataset. In our proposed strategy, where we use adversarial watermarks, we can see that it outperforms the baseline with respect to its robustness against evasion attacks. In terms of its adversarial accuracy, it has less than 1\% drop (from 92.82\% $\rightarrow$ 92.01\%) for the MNIST dataset and around 4\% drop (from 70.95\% $\rightarrow$ 65.84\%) for the FMNIST dataset. We attribute this slight decrease in robustness performance to an unintentional conflict that might arise concerning the interplay between the perturbation budget used to craft the adversarial samples and watermarks. However, the overall results obtained empirically support our hypothesis of using the watermarks with a similar distribution as that of adversarial samples to enhance the robustness of the model, while also maintaining comparable performance in terms of test and watermarking accuracy.  
% Moreover, one interesting finding can be seen in case of watermarking accuracy (for proposed technique), where it is degraded from 99\% to 90\% in case on FMNIST dataset. We claim this is due to the perturbation 
\begin{figure}[h!]

\subfloat[Pruning Attack]{
  \includegraphics[width=\columnwidth]{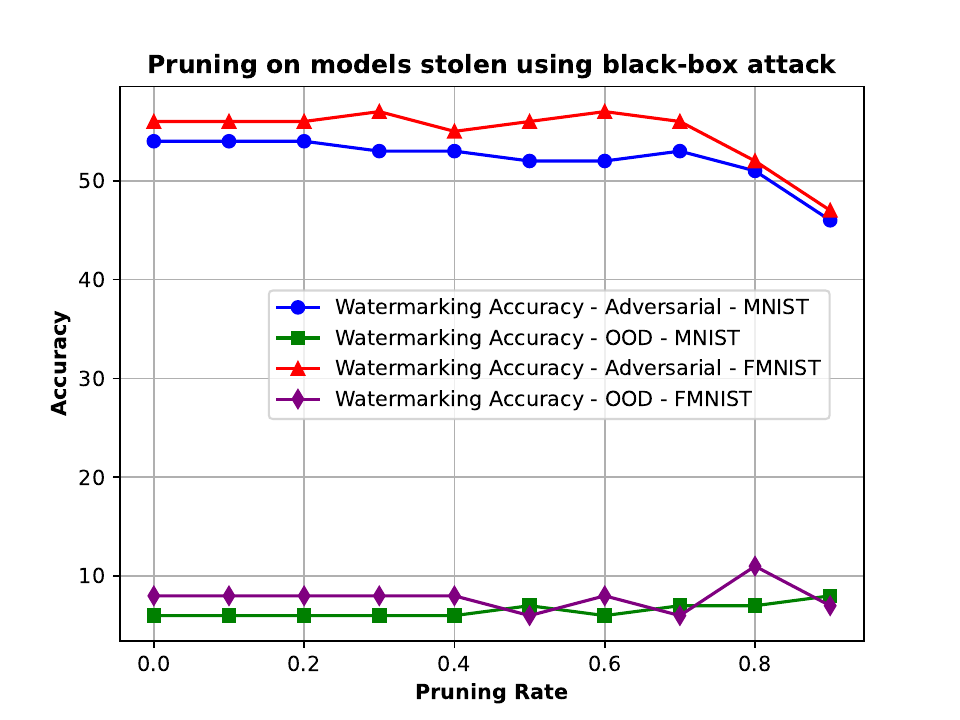}%
  % \label{black_prune}
}

\subfloat[Fine-tuning Attack]{
  \includegraphics[width=\columnwidth]{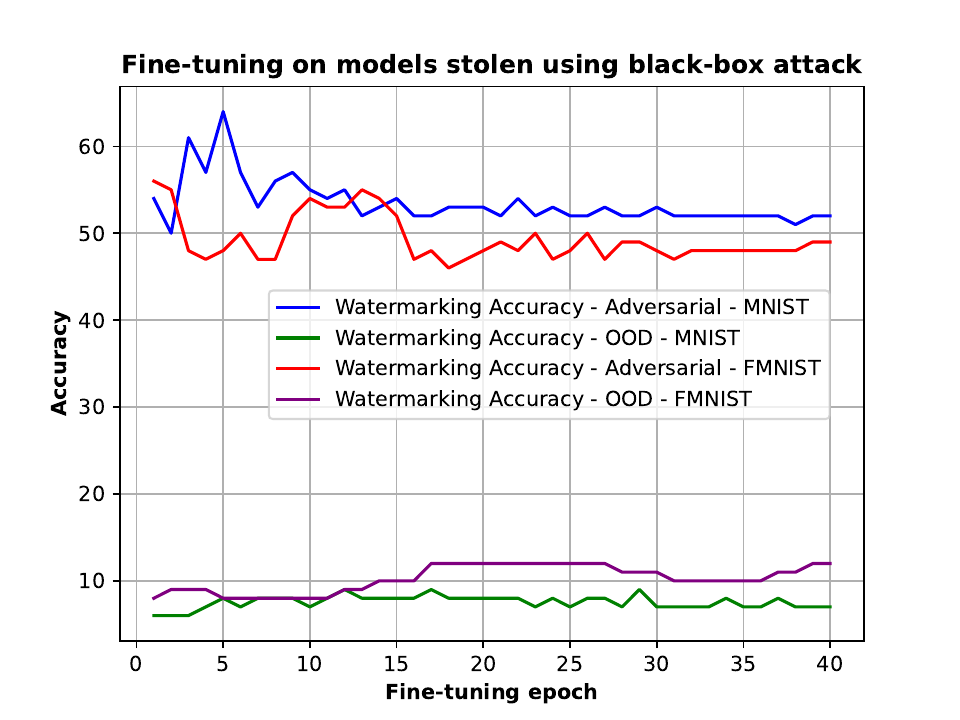}%
}%
\caption{Impact of removal attack on the model stolen using black-box setting}
\label{fig:black-box}
\end{figure}
\begin{table*}[h!]
\centering
\begin{tabular}{c ccc ccc}
\toprule
\multicolumn{7}{c}{\textbf{Grey-box Transferability}} \\ \toprule
\multicolumn{1}{c}{} &
  \multicolumn{3}{c}{\textbf{AdvTraining + WM (OOD)}} &
  \multicolumn{3}{c}{\textbf{AdvTraining + WM (Adversarial)}} \\
\multicolumn{1}{c}{\multirow{-2}{*}{\textbf{Dataset}}} &
  \multicolumn{1}{c}{\textbf{Test Acc}} &
  \multicolumn{1}{c}{\textbf{Adv Acc}} &
  \multicolumn{1}{c}{\textbf{Water Acc}} &
  \multicolumn{1}{c}{\textbf{Test Acc}} &
  \multicolumn{1}{c}{\textbf{Adv Acc}} &
  \textbf{Water Acc} \\ \midrule
\multicolumn{1}{c|}{\textbf{MNIST}} &
  \multicolumn{1}{c|}{91.23} &
  \multicolumn{1}{c|}{46.3} &
  \multicolumn{1}{c|}{9} &
  \multicolumn{1}{c|}{91.03} &
  \multicolumn{1}{c|}{45.29} &
  68 \\
\multicolumn{1}{c|}{\textbf{FMNIST}} &
  \multicolumn{1}{c|}{55.61} &
  \multicolumn{1}{c|}{41.58} &
  \multicolumn{1}{c|}{5} &
  \multicolumn{1}{c|}{75.33} &
  \multicolumn{1}{c|}{42.85} &
  60 \\ \bottomrule
\end{tabular}
\caption{Transferability of performance for various metric when the models undergo grey-box model stealing attack.}
\label{tab:grey-box-performance}
\end{table*}
\subsection{Robustness in Black-box Setting}
In this section, we examine the transferability of our watermarks to the model stolen using the black-box model stealing attack. As we can observe in Table \ref{tab:black-box-performance}, test accuracy is high for both datasets for both approaches\footnote{The test accuracy is quite close to the non-stolen original model in a real-world setting, as evident in the literature \cite{papernot2017practical}. However, due to the resource limitation, it is challenging to launch strong model stealing attacks for our research, thus resulting in a noticeable further decline in performance.}. Moreover, one can notice that transferability for adversarial samples for both approaches is very low. We claim this is because, while launching a black-box attack, we had no information that the model was trained using adversarial training. Our adversary only queried the pure input-output pairs, thus limiting our model performance on the adversarial samples. However, we can notice a high transferability of our watermarks for the model which was trained using adversarial watermarks. 

Furthermore, one can observe that even if the model was trained simultaneously using adversarial training, the adversarial watermarks did not conflict with the adversarial samples and were independently verified with high confidence. The results confirms our understanding regarding the conflict between adversarial watermarks and adversarial samples, and demonstrates the efficacy of our suggested interplay of the two techniques in a black-box setting. 
\subsubsection{With respect to Pruning}
Figure \ref{fig:black-box}a, plots the effect of pruning the stolen model with different pruning rates for two datasets. From Table \ref{tab:black-box-performance}, we know that the transferability of the OOD dataset is quite low, and thus, applying further removal attacks does not significantly affect its behavior. Further, one can observe that, with increasing pruning rate, our approach can still verify the ownership of the model with high confidence, i.e., for both the datasets, the models can be confidently verified with more than 50\% transferability rate, with as high as 80\% pruned neurons. This implies that the embedded watermarks significantly contribute to the important neurons of our model. Thus, they cannot be easily removed without degrading the model performance. 

\subsubsection{With respect to Fine-tuning}
Figure \ref{fig:black-box}b plots the effect of fine-tuning the stolen model with 40 epochs. The OOD watermarking accuracy by default is low due to its low transferability (Table \ref{tab:black-box-performance}). Thus fine-tuning it further does not give us any useful information. However, we can see that when we fine-tune the models trained using adversarial watermarks, the watermarking accuracy decreases to a certain point and then nearly stays constant throughout the remaining process. Although we observe a decrease in watermarking accuracy, it is still high enough (more than 45\% for both datasets) to confidently verify the ownership of the models. The findings empirically show the effectiveness of our approach to pruning and fine-tuning attacks in the black-box setting.

\begin{figure}[h!]

\subfloat[Pruning Attack]{
  \includegraphics[width=\columnwidth]{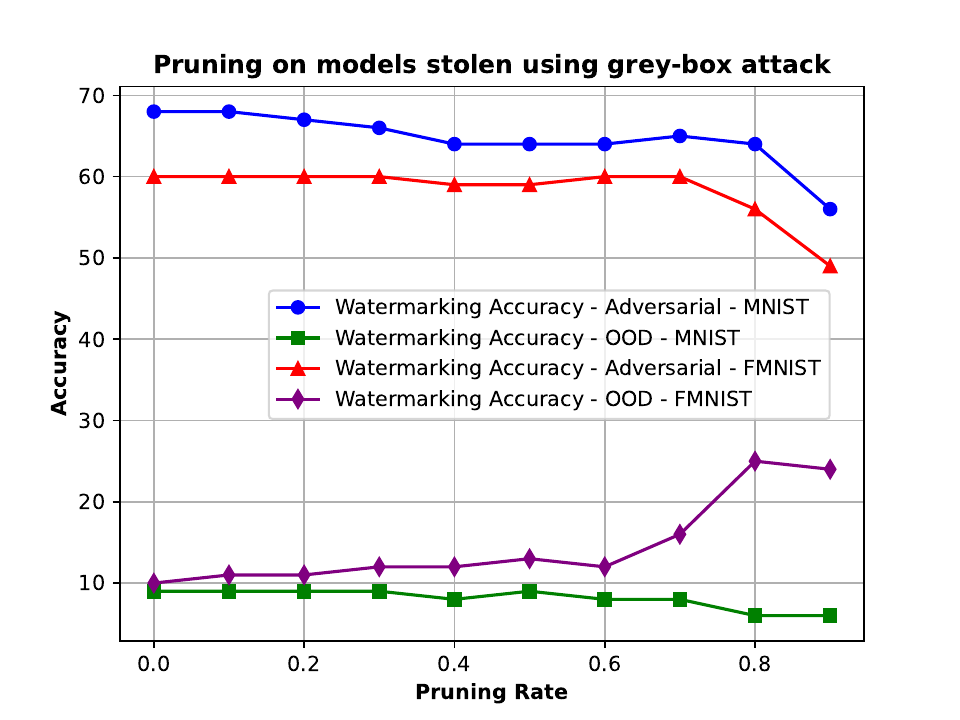}%
  % \label{black_prune}
}

\subfloat[Fine-tuning Attack]{
  \includegraphics[width=\columnwidth]{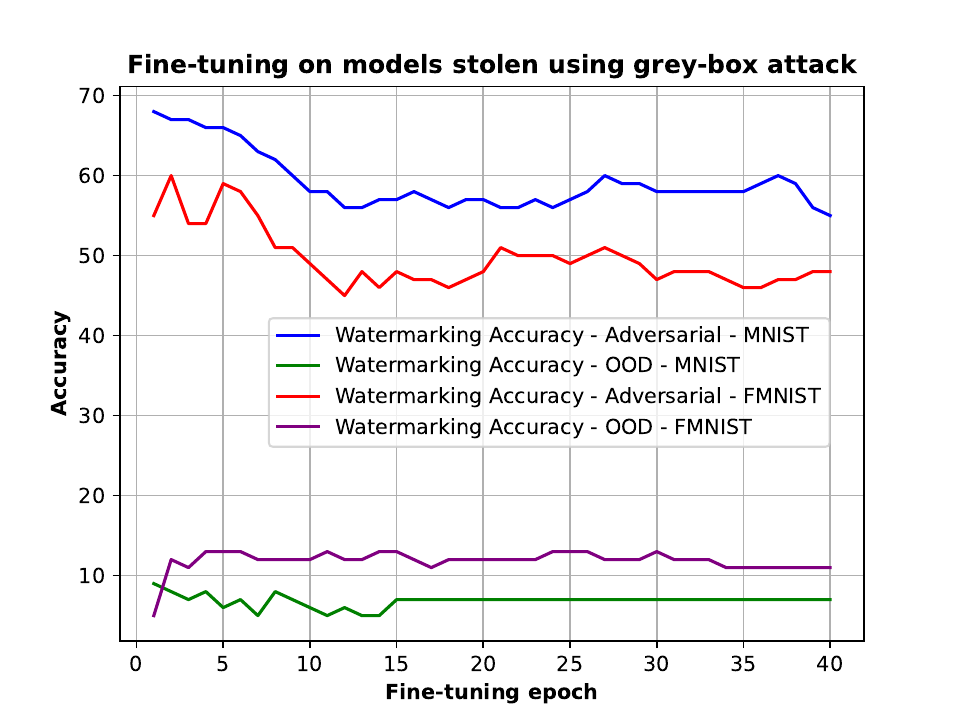}%
}%
\caption{Impact of removal attack on the model stolen using grey-box setting}
\label{fig:grey-box}
\end{figure}

\subsection{Robustness in Grey-box Setting}
In this section we investigate the transferability of our watermarks to the model stolen using the grey-box model stealing attack. In comparison to the black-box scenario, in this setting, we have information about the architecture of the model and the fact that it was trained using adversarial training. As shown in Table \ref{tab:grey-box-performance}, test accuracy is high for both approaches. The use of additional information regarding adversarial training helped to launch a more powerful attack, thus making the stolen model perform well on adversarial samples. However, with additional information comes a risk, as one can notice that the confidence with which the watermarks can be detected has increased in the grey-box setting (compared to the black-box) for our proposed strategy. So, even if an adversary launches a powerful model stealing attack with more information, it increases the transferability of our adversarial watermarks, thus putting the adversaries at high risk of getting caught.

\subsubsection{With respect to Pruning}
Figure \ref{fig:grey-box}a, plots the effect of pruning the stolen model using a grey-box attack for two datasets. Similar to the case of black-box setting, the watermarking accuracy for the model with the OOD dataset is quite low (Table \ref{tab:grey-box-performance}). Thus, applying a pruning attack on it does not provide any further insights. However, our proposed strategy performs much better compared to the baseline and can confidently verify the ownership of the model. One can further notice that the detection rate for both the datasets is still high when 80\% of the neurons are pruned, i.e., the models can be confidently verified with more than 55\% detection rate. 

\subsubsection{With respect to Fine-tuning}
Figure \ref{fig:grey-box}b, plots the effect of fine-tuning the stolen model with a total of 40 epochs. Similar to our observation in case of black-box setting, we can see that when we fine-tune the models trained using adversarial watermarks, the watermarking accuracy drops up to a particular point and then remains nearly constant throughout the tuning process. The findings above empirically show the effectiveness of our approach to pruning and fine-tuning attacks in the grey-box setting. 

\begin{figure}[h!]

\subfloat[Pruning Attack]{
  \includegraphics[width=\columnwidth]{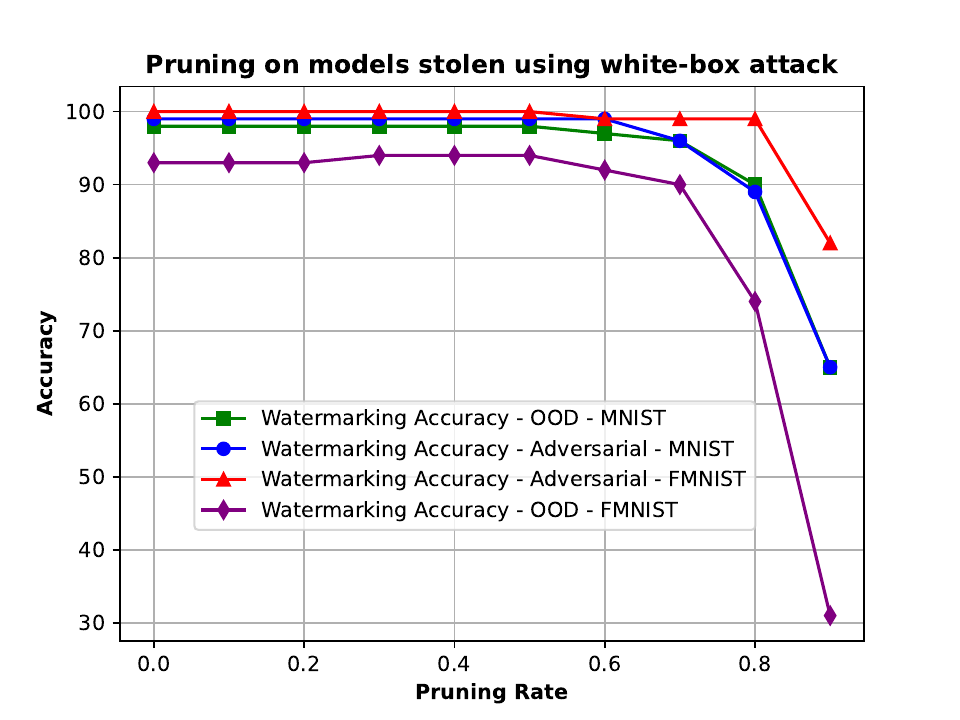}%
  % \label{black_prune}
}

\subfloat[Fine-tuning Attack]{
  \includegraphics[width=\columnwidth]{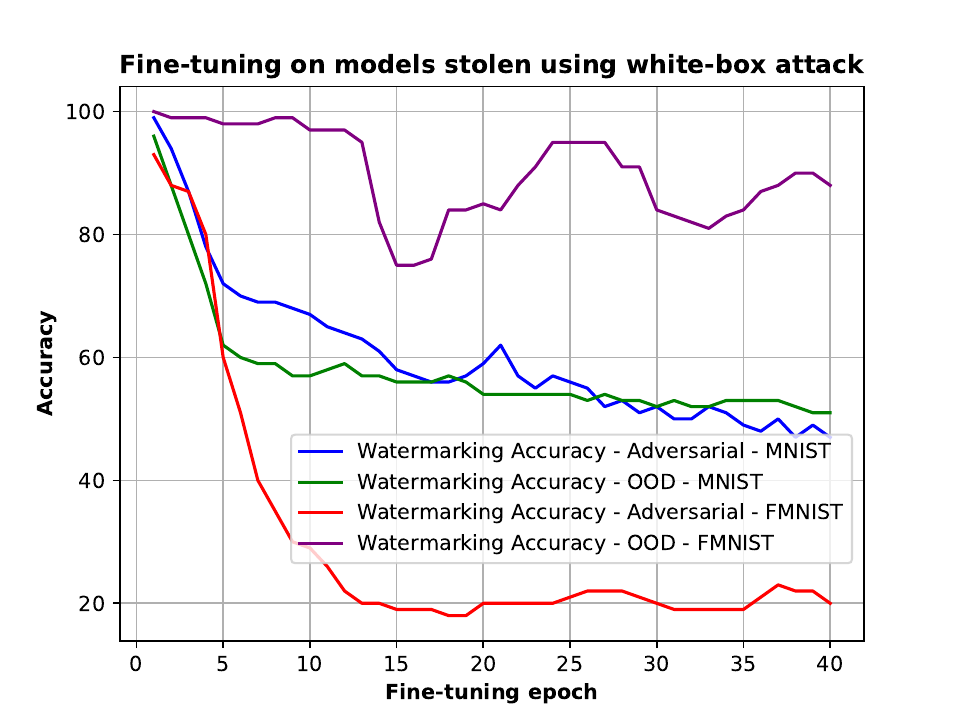}%
}%
\caption{Impact of removal attack on the model stolen using white-box setting}
\label{fig:white-box}
\end{figure}
\subsection{Robustness in White-box Setting}
In this section, we observe the transferability of our watermarks to the model stolen using a white-box model stealing attack. This particular scenario is specific to having full access to the model. So, for this experiment, we directly took the original model to carry out the analysis. The performance of the model stolen using a white-box scenario would be exactly the same as the non-stolen model (Table \ref{tab:performance2}). For the white-box setting, we are more interested in watermarks' resistance to removal attacks. 

\subsubsection{With respect to Pruning}
Figure \ref{fig:white-box}a plots the effect of pruning the stolen model using a white-box attack. The transferability for the watermarks is quite high for both OOD and adversarial watermarks. However, with an increase in pruning rate, when more than 70\% of the neurons are pruned, there is a significant drop in the watermarking accuracy. For the case of the MNIST dataset (for both approaches), it drops to 65\% when the pruning rate is 0.9. However, we can still confidently verify its ownership with high confidence. In the case of the FMNIST dataset, we can see that for the model trained using adversarial watermarks, the accuracy drops to almost 30\% when the pruning rate is 0.9, while the approach with OOD watermarks outperforms our proposed technique in this particular scenario. We attribute this to the complexity of the FMNIST dataset, which hinders the model from fully overfitting, resulting into a drop in accuracy when pruned heavily. This observation is only specific to white-box attacks on FMNIST; we believe that in white-box attacks, the information learned is distributed among the neurons, and the model might not entirely focus on overfitting to watermarks. Conversely, in black-box and grey-box settings, we query the cloud API to generate the input-output pairs, which forces the transfer of the watermarks with their performance. 

\subsubsection{With respect to Fine-tuning}
Figure \ref{fig:white-box}b illustrates the effect of fine-tuning the stolen model with 40 epochs. The initial transferability rate for both baseline and proposed strategy is quite high. However, for both datasets, we observe that when the fine-tuning epoch increases, it starts to lose its resilience towards watermarks. In particular, the OOD watermark accuracy observes a high drop in the case of the MNIST dataset, which we believe is due to the use of the FMNIST dataset as OOD watermarks, which is complex and thus makes it challenging to overfit the models on these watermarks. In contrast, the OOD watermarking accuracy for the FMNIST dataset, which uses the MNIST dataset as OOD watermarks, remains quite high. Moreover, the drop in accuracy for the models trained using adversarial watermarks is higher than the baseline approach for the FMNIST dataset. This is similar to the trend we observed in the pruning attacks. We believe this is due to the complexity of the FMNIST dataset and the difference between how the model is stolen using various model stealing attacks.

After thoroughly evaluating the performance of our proposed strategy and baseline approach on various model stealing and removal attacks, we observed that our proposed technique of using adversarial watermarks combined with adversarial training consistently outperforms the baseline in nearly all scenarios, with the exception of the white-box attack on the FMNIST dataset. Although we acknowledge this limitation, it is important to highlight that our approach exhibits remarkable superiority in grey-box and black-box settings. The key benefit of our proposed approach is its high transferability in grey-box and black-box scenarios, which are more practical and commonly occurring attacks in real world compared to rare white-box attacks. Thus, the efficacy of our technique against such attacks further validates its utility in the real world. 

\section{Conclusions}
In conclusion, this study introduced a novel way of combining adversarial watermarks and adversarial training without undermining its primary objectives. We observed that there exists a lower bound perturbation budget above which the utility of the model worsens, making it ineffective. We leverage this information to generate the adversarial watermarks that differ from the adversarial samples used in the training. We benchmark the performance of our strategy on various model stealing and removal attacks. Our proposed technique consistently outperforms the baseline in nearly all scenarios. 

\section{Acknowledgement}

This work was supported in part by the European Union's Horizon 2020 research and innovation programme under grant number 951911 – AI4Media.

\bibliography{aaai24}

\begin{thebibliography}{15}
\providecommand{\natexlab}[1]{#1}

\bibitem[{Deng(2012)}]{deng2012mnist}
Deng, L. 2012.
\newblock The mnist database of handwritten digit images for machine learning research [best of the web].
\newblock \emph{IEEE signal processing magazine}, 29(6): 141--142.

\bibitem[{Goodfellow, Shlens, and Szegedy(2014)}]{goodfellow2014explaining}
Goodfellow, I.~J.; Shlens, J.; and Szegedy, C. 2014.
\newblock Explaining and harnessing adversarial examples.
\newblock \emph{arXiv preprint arXiv:1412.6572}.

\bibitem[{Kurakin, Goodfellow, and Bengio(2018)}]{kurakin2018adversarial}
Kurakin, A.; Goodfellow, I.~J.; and Bengio, S. 2018.
\newblock Adversarial examples in the physical world.
\newblock In \emph{Artificial intelligence safety and security}, 99--112. Chapman and Hall/CRC.

\bibitem[{Le~Merrer, Perez, and Tr{\'e}dan(2020)}]{le2020adversarial}
Le~Merrer, E.; Perez, P.; and Tr{\'e}dan, G. 2020.
\newblock Adversarial frontier stitching for remote neural network watermarking.
\newblock \emph{Neural Computing and Applications}, 32: 9233--9244.

\bibitem[{Li et~al.(2022)Li, Fan, Gu, Li, and Yang}]{li2022fedipr}
Li, B.; Fan, L.; Gu, H.; Li, J.; and Yang, Q. 2022.
\newblock FedIPR: Ownership verification for federated deep neural network models.
\newblock \emph{IEEE Transactions on Pattern Analysis and Machine Intelligence}.

\bibitem[{Madry et~al.(2017)Madry, Makelov, Schmidt, Tsipras, and Vladu}]{madry2017towards}
Madry, A.; Makelov, A.; Schmidt, L.; Tsipras, D.; and Vladu, A. 2017.
\newblock Towards deep learning models resistant to adversarial attacks.
\newblock \emph{arXiv preprint arXiv:1706.06083}.

\bibitem[{Namba and Sakuma(2019)}]{namba2019robust}
Namba, R.; and Sakuma, J. 2019.
\newblock Robust watermarking of neural network with exponential weighting.
\newblock In \emph{Proceedings of the 2019 ACM Asia Conference on Computer and Communications Security}, 228--240.

\bibitem[{Papernot et~al.(2017)Papernot, McDaniel, Goodfellow, Jha, Celik, and Swami}]{papernot2017practical}
Papernot, N.; McDaniel, P.; Goodfellow, I.; Jha, S.; Celik, Z.~B.; and Swami, A. 2017.
\newblock Practical black-box attacks against machine learning.
\newblock In \emph{Proceedings of the 2017 ACM on Asia conference on computer and communications security}, 506--519.

\bibitem[{Szegedy et~al.(2013)Szegedy, Zaremba, Sutskever, Bruna, Erhan, Goodfellow, and Fergus}]{szegedy2013intriguing}
Szegedy, C.; Zaremba, W.; Sutskever, I.; Bruna, J.; Erhan, D.; Goodfellow, I.; and Fergus, R. 2013.
\newblock Intriguing properties of neural networks.
\newblock \emph{arXiv preprint arXiv:1312.6199}.

\bibitem[{Szyller and Asokan(2022)}]{szyller2022conflicting}
Szyller, S.; and Asokan, N. 2022.
\newblock Conflicting Interactions Among Protections Mechanisms for Machine Learning Models.
\newblock \emph{arXiv preprint arXiv:2207.01991}.

\bibitem[{Szyller et~al.(2021)Szyller, Atli, Marchal, and Asokan}]{szyller2021dawn}
Szyller, S.; Atli, B.~G.; Marchal, S.; and Asokan, N. 2021.
\newblock Dawn: Dynamic adversarial watermarking of neural networks.
\newblock In \emph{Proceedings of the 29th ACM International Conference on Multimedia}, 4417--4425.

\bibitem[{Tram{\`e}r et~al.(2017)Tram{\`e}r, Kurakin, Papernot, Goodfellow, Boneh, and McDaniel}]{tramer2017ensemble}
Tram{\`e}r, F.; Kurakin, A.; Papernot, N.; Goodfellow, I.; Boneh, D.; and McDaniel, P. 2017.
\newblock Ensemble adversarial training: Attacks and defenses.
\newblock \emph{arXiv preprint arXiv:1705.07204}.

\bibitem[{Uchida et~al.(2017)Uchida, Nagai, Sakazawa, and Satoh}]{uchida2017embedding}
Uchida, Y.; Nagai, Y.; Sakazawa, S.; and Satoh, S. 2017.
\newblock Embedding watermarks into deep neural networks.
\newblock In \emph{Proceedings of the 2017 ACM on international conference on multimedia retrieval}, 269--277.

\bibitem[{Xiao, Rasul, and Vollgraf(2017)}]{xiao2017fashion}
Xiao, H.; Rasul, K.; and Vollgraf, R. 2017.
\newblock Fashion-mnist: a novel image dataset for benchmarking machine learning algorithms.
\newblock \emph{arXiv preprint arXiv:1708.07747}.

\bibitem[{Zhu and Gupta(2017)}]{zhu2017prune}
Zhu, M.; and Gupta, S. 2017.
\newblock To prune, or not to prune: exploring the efficacy of pruning for model compression.
\newblock \emph{arXiv preprint arXiv:1710.01878}.

\end{thebibliography}

\onecolumn
\section*{Appendix}\label{label:appendix}
\begin{algorithm}[]
\DontPrintSemicolon
  
  \KwInput{\\
  $\textbf{$X_{training}$} \leftarrow \text{datapoints}\{(x_1,y_1),(x_2,y_2), ....,(x_N,y_N)\}$ \\
  $\textbf{$X_{watermarks}$} \leftarrow \text{datapoints}\{(x_{w1},y_{w1}),(x_{w2},y_{w2}), ...., (x_{wN},y_{wN})\}$ (\textbf{OOD or Adversarial})\\
   \textbf{Loss function}: $L(\theta) \leftarrow \frac{1}{N} \sum_{i=1}^N L(\theta, x_i)$\\
  \textbf{Hyperparameters:} \\
  \textbf{Learning rate}: $\alpha$, \textbf{Adversarial Learning rate:} $\beta$, \textbf{Iterations:}\text{ $T$}, \textbf{batch size:} \text{ $B$}, \textbf{Perturbation budget:} $\epsilon$, \textbf{Attack steps:} \text{ $attack\_steps$} } 
% \textbf{Clipping threshold} (gradient norm bound): $C$\\
% \textbf{Number of mini-batches:} $mB = \frac{N}{B}$}
% Select $k$ centroids $\textbf{S}^{(0)} = ({S}^{(0)}_{1},{S}^{(0)}_{2}, ..., {S}^{(0)}_{k})$ uniformly from X.
% \\
% $iterationForLloyd$ = number of iterations to run the algorithm.
% \\
 \For{$t \in [T]$}{
    
    \For{\text{each i} $\in X_{training\_batch}$}{
        \textbf{Generate the Adversarial Samples}\\
        $B_{adv\_i} \leftarrow \text{PGD}(B_i, \epsilon, attack\_steps, \beta)$ \\
        \textbf{Compute the gradients}\\
         \text{compute} $g_t (B_{adv\_i}) \leftarrow \nabla_{\theta_t}\mathcal{L}(\theta_t, B_{adv\_i})$\\
         % \textbf{Clip Gradient}\\
         % $\overline{g}_t(x_i) \gets \frac{g_t(x_i)}{\max \left(1, \frac{\lVert g_t(x_i) \rVert_2}{C}\right)}$\\
         % \textbf{Add Noise}\\
         %  $\tilde{g}_t \gets \frac{1}{B} \sum_{i}\left(\overline{g}_t(x_i) + \mathcal{N}(0, \sigma^2C^2I)\right)$\\
          \textbf{Gradient descent step}\\
          $\theta_{t+1} \gets \theta_t - \alpha g_t (B_{adv\_i})$\\
        % ${{C}^{(t)}_{i}} \leftarrow$ assign each $x_j$ to its closest centroid ${S_i}^{t-1}$; \\
        % ${S_i}^{t} \leftarrow$ centroid of ${C_i}^{t}$;\\
        % ${ConvergentZone_i}^{(t)} \leftarrow$ List of data points inside the spherical region having ${S_i}^{t}$ and ${S_i}^{t-1}$ as the endpoints of its radius.\\
        % ${SamplingZone_i}^{(t)} \leftarrow$ run Algorithm 2 using ${ConvergentZone_i}^{(t)}$ , $internalK$;\\
        % $\hat{S_i}^{(t)} \leftarrow$ sample from ${SamplingZone_i}^{(t)}$ using ExpDP with $q$ and $\epsilon^{exp}$;\\
        % ${S_i}^{(t)} \leftarrow$ $\hat{S_i}^{(t)}$
    }
    \For{\text{each i} $\in X_{watermarks\_batch}$}{
        % \textbf{Generate the Adversarial Samples}\\
        % $B_{adv\_i} = \text{PGD}(B_i, \epsilon, attack\_steps, \beta)$ \\
        \textbf{Compute the gradients}\\
         \text{compute} $g_t (B_i)) \leftarrow \nabla_{\theta_t}\mathcal{L}(\theta_t, B_i)$\\
         % \textbf{Clip Gradient}\\
         % $\overline{g}_t(x_i) \gets \frac{g_t(x_i)}{\max \left(1, \frac{\lVert g_t(x_i) \rVert_2}{C}\right)}$\\
         % \textbf{Add Noise}\\
         %  $\tilde{g}_t \gets \frac{1}{B} \sum_{i}\left(\overline{g}_t(x_i) + \mathcal{N}(0, \sigma^2C^2I)\right)$\\
          \textbf{Gradient descent step}\\
          $\theta_{t+1} \gets \theta_t - \alpha g_t (B_i)$\\
    }
 }
 \KwOutput{$\theta_{T}$}

\caption{Adversarial Training and Model Watermarking}
\label{adv_water_algo}
\end{algorithm}

\begin{figure}[h!]
\centering
\includegraphics[width = \linewidth]{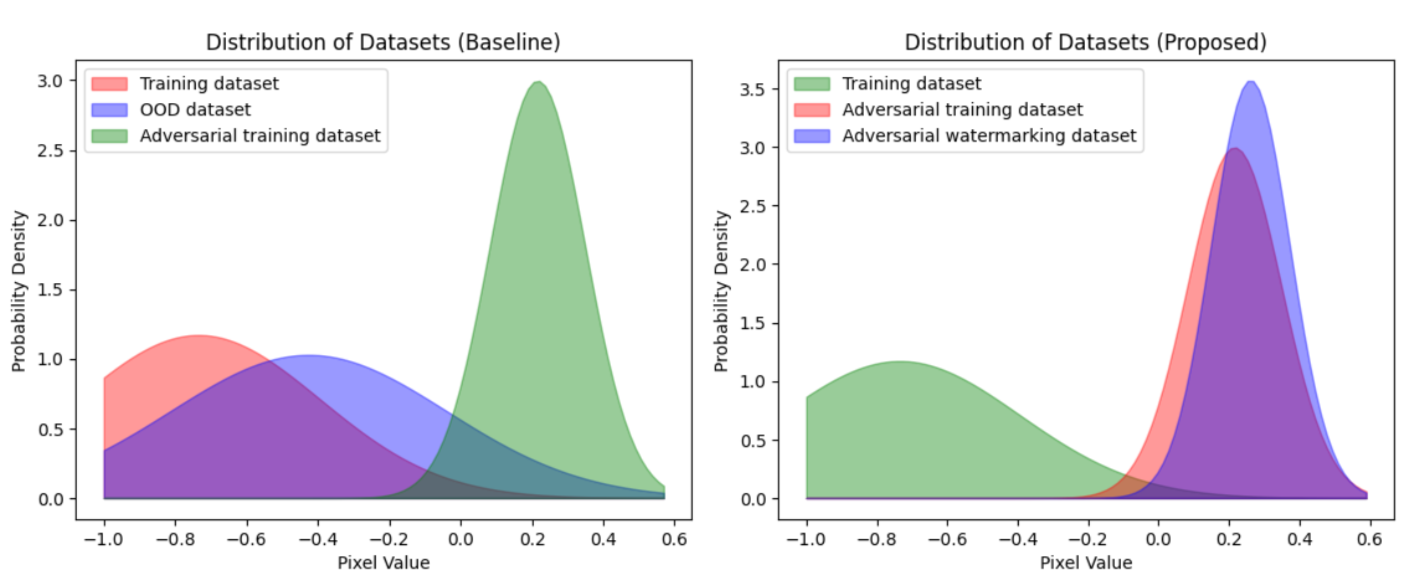}
\caption{Above figure plots the distribution of datasets for MNIST dataset, including its training dataset, adversarial training dataset and watermarking set (OOD - for baseline, and adversarial watermarks for the proposed use case). The OOD dataset used for the baseline is Fashion-MNIST dataset.}
\label{fig:samples}
\end{figure}
% \usepackage{multirow}
\begin{table}[h!]
\centering
\begin{tabular}{|c|cr|crrrr|}
\hline
\multirow{2}{*}{\textbf{Dataset}} & \multicolumn{2}{c|}{\textbf{Adversarial Training}}                             & \multicolumn{5}{c|}{\textbf{Adversarial Accuracy (perturbation budget, attack steps)}}                                                                                                                       \\
                                  & \multicolumn{1}{c|}{\textbf{Test Acc}} & \multicolumn{1}{c|}{\textbf{Adv Acc}} & \multicolumn{1}{c|}{\textbf{(0.15,15)}} & \multicolumn{1}{c|}{\textbf{(0.25,25)}} & \multicolumn{1}{c|}{\textbf{(0.3,40)}} & \multicolumn{1}{c|}{\textbf{(0.4,40)}} & \multicolumn{1}{c|}{\textbf{(0.5,40)}} \\ \hline
\textbf{MNIST}                    & \multicolumn{1}{c|}{99.2}              & 92.82                                 & \multicolumn{1}{c|}{96.34}              & \multicolumn{1}{c|}{92.82}              & \multicolumn{1}{c|}{46.28}             & \multicolumn{1}{c|}{0.1}               & 0                                      \\ 
\textbf{FMNIST}                   & \multicolumn{1}{c|}{86.91}             & 70.95                                 & \multicolumn{1}{c|}{70.95}              & \multicolumn{1}{c|}{5.46}               & \multicolumn{1}{c|}{0.43}              & \multicolumn{1}{c|}{0}                 & 0                                      \\ \hline
\end{tabular}
\caption{Adversarial Training and its corresponding robustness to adversarial samples generated using different perturbation budgets.}
\label{only-adv-1}
\end{table}
\subsection{Hyperparameter}
% \usepackage{multirow}
\begin{table}[h!]
\centering
\begin{tabular}{|c|ccc|ccc|c|c|}
\hline
\multirow{2}{*}{Dataset} & \multicolumn{3}{c|}{ADVTR}                                                          & \multicolumn{3}{c|}{ADVWM}                                                        & \multirow{2}{*}{\begin{tabular}[c]{@{}c@{}}Watermark\\ $| D_{W}|$\end{tabular}} & \multicolumn{1}{c|}{\multirow{2}{*}{OOD dataset}} \\
                         & \multicolumn{1}{c|}{$\epsilon$} & \multicolumn{1}{c|}{$\gamma$} & \multicolumn{1}{c|}{$\eta$} & \multicolumn{1}{c|}{$\epsilon$} & \multicolumn{1}{c|}{$\gamma$} & \multicolumn{1}{c|}{$\eta$} &                                                                                   & \multicolumn{1}{c|}{}                             \\ \hline
MNIST                    & \multicolumn{1}{r|}{0.25}    & \multicolumn{1}{r|}{25}           & 0.01                            & \multicolumn{1}{r|}{0.4}     & \multicolumn{1}{r|}{40}           & 0.01                            & \multicolumn{1}{c|}{100}                                                          & Fashion-MNIST                                     \\
FMNIST                   & \multicolumn{1}{r|}{0.15}    & \multicolumn{1}{r|}{15}           & 0.01                            & \multicolumn{1}{r|}{0.3}     & \multicolumn{1}{r|}{40}           & 0.01                            & \multicolumn{1}{c|}{100}                                                          & MNIST                                             \\ \hline
\end{tabular}
\caption{Hyperparameter for MNIST and Fashion-MNIST dataset}
\label{hyper-params}
\end{table}
We used the hyperparameters defined in the \textbf{Table: \ref{hyper-params}} for our experiments. The ($\epsilon$, $\gamma$, $\eta$) stands for (epsilon budget, attack steps, step size), respectively, for adversarial training (ADVTR) and adversarial watermarks (ADVWM). In addition, all the models were trained for a total of $100$ epochs, with a learning rate of $0.005$ and weight decay of $5e-4$. As a standard training setting, we utilized the SGD optimizer to iteratively update the model parameters. Please note that while selecting the OOD dataset, there is no direct correlation in using watermarks as FMNIST for the MNIST dataset or vice-versa. One can choose any arbitrary data points as their watermarks as long as they are out-of-distribution with respect to the training dataset. 

\begin{algorithm}[]
\DontPrintSemicolon
  
  \KwInput{$\mathcal{O} \leftarrow$ \text{Oracle model or target model (API access)} \\
  $epoch \leftarrow$ \text{substitute model training epochs} \\
  $S_{train} \leftarrow$ \text{initial training set} \\
  $F_{BLACK} \leftarrow$ \text{architecture for black-box attack}\\
  $F_{GREY} \leftarrow$ \text{architecture for grey-box attack}
  } 
% \\
Define architecture $F_{BLACK}$ or $F_{GREY}$ based on the attack \\
 \For{$i \in epoch$}{
    \textbf{Query the substitute training set}\\
    $\mathcal{D} \leftarrow \{ (x, \mathcal{O}(x)): x \in S_{train_{i}}\}$\\
    \textbf{Train F on $\mathcal{D}$ to evaluate parameters $\theta_{F}$} \\
    \If{\textit{black-box attack}}{
     $\theta_{F_{BLACK}} \leftarrow$ train($F_{BLACK},\mathcal{D}$)
    }
    \Else{
    $\theta_{F_{GREY}} \leftarrow$ \text{train($F_{GREY},\mathcal{D}$)} // \text{adversarial training}
    }
    \textbf{Perform Jacobian-based dataset augmentation} \\
    $S_{train_{i+1}} \leftarrow$ $\{ x + \lambda \cdot sgn(J_{F} [\mathcal{O}(x)]): x \in S_{train_{i}}\} \cup S_{train_{i}}$ \\
 }
 \KwOutput{$\theta$}

\caption{Model Stealing Attack \cite{papernot2017practical}}
\label{model_steal}
\end{algorithm}
\begin{algorithm}
  \caption{Removal Attack \cite{li2022fedipr}}
  \SetAlgoLined
  \KwInput{$\mathcal{M} \leftarrow$ \text{stolen model,} \\ 
  $p \leftarrow$ \text{pruning-rate, } $D_{\text{add}} \leftarrow$ \text{additional training data}
  }
  \SetKwProg{proc}{procedure}{:}{end}
  \proc{PRUNING}{
    Prune the model $N$ with pruning rate $p$.
  }

  \SetKwProg{proc}{procedure}{:}{end}
  \proc{FINETUNING}{
    \For{epochs in 50}{
      Train the model $N$ only on the main classification task with additional training data $D_{\text{add}}$.
    }
  }
\label{removal}
\end{algorithm}

\end{document}